\documentclass[letterpaper, 10 pt, conference]{ieeeconf}  

\pdfoutput=1

\IEEEoverridecommandlockouts                              

\usepackage{graphics} 
\usepackage{graphicx}

\usepackage{epsfig} 
\usepackage{amsmath} 
\usepackage{amssymb}  

\usepackage{bm}
\usepackage{mathtools}

\usepackage{citesort}
\let\labelindent\relax
\usepackage{enumitem}

\usepackage{algorithm}
\usepackage[noend]{algpseudocode} 
\algnewcommand\AAND{\textbf{ and }}
\algnewcommand\Or{\textbf{ or }}

\usepackage{color}
\usepackage{flushend}
\usepackage{url}
\usepackage[breaklinks]{hyperref}
\usepackage[capitalize]{cleveref}
\usepackage{booktabs}
\usepackage{makecell}
\usepackage{multirow}

\usepackage[nolist,nohyperlinks]{acronym}
\acrodef{method}[DR-LRIO]{Degradation Resilient LiDAR-Radar-Inertial Odometry}
\acrodef{gnss}[GNSS]{Global Navigation Satellite System}
\acrodef{icp}[ICP]{Iterative Closest Point}
\acrodef{uas}[UAS]{Unmanned Aerial Systems}
\acrodef{ransac}[RANSAC]{Random Sample Consensus}
\acrodef{slam}[SLAM]{Simultaneous Localization And Mapping}
\acrodef{fmcw}[FMCW]{Frequency Modulated Continuous Wave}
\acrodef{pca}[PCA]{Principal Component Analysis}
\acrodef{ekf}[EKF]{Extended Kalman Filter}
\acrodef{isam2}[iSAM2]{Incremental Smoothing and Mapping}
\acrodef{rio}[RIO]{Radar Inertial Odometry}
\acrodef{rmse}[RMSE]{Root Mean Square Error} 
\acrodef{ape}[APE]{Absolute Pose Error}
\acrodef{cfar}[CFAR]{Constant False Alarm Rate}
\acrodef{snr}[SNR]{Signal to Noise Ratio}
\acrodef{rcs}[RCS]{Radar Cross Section}
\acrodef{imu}[IMU]{Inertial Measurement Unit}

\usepackage{tikz}
\usetikzlibrary{positioning, shapes.geometric, arrows.meta, decorations.pathreplacing, calligraphy}

\DeclareMathAlphabet{\pazocal}{OMS}{zplm}{m}{n}

\DeclareMathOperator*{\argmin}{argmin}
\DeclareMathOperator*{\argmax}{argmax} 

\DeclareMathAlphabet{\mathpzc}{OT1}{pzc}{m}{it}

\usepackage{array}
\newcolumntype{C}[1]{>{\centering\arraybackslash}p{#1}}
\newcolumntype{M}[1]{>{\raggedright\arraybackslash}p{#1}}

\usepackage{array} 
\newcolumntype{L}[1]{>{\raggedright\let\newline\\\arraybackslash\hspace{0pt}}m{#1}}	
\newcolumntype{S}[1]{>{\centering\let\newline\\\arraybackslash\hspace{0pt}}m{#1}}
\newcolumntype{R}[1]{>{\raggedleft\let\newline\\\arraybackslash\hspace{0pt}}m{#1}}

\makeatletter
\renewcommand*{\@opargbegintheorem}[3]{\trivlist
  \item[\hskip \labelsep{\itshape #1\ #2}] \textit{(#3)}\ }
\makeatother

\usepackage[per-mode=symbol]{siunitx}

\title{\LARGE \bf
Degradation Resilient LiDAR-Radar-Inertial Odometry
}

\author{Morten Nissov$^\star$, Nikhil Khedekar$^\star$, and Kostas Alexis 
\thanks{$^\star$ The authors contributed equally.} 
\thanks{This material was supported by the Research Council of Norway Award NO-321435.}
\thanks{The authors are with the Norwegian University of Science and Technology (NTNU), O. S. Bragstads Plass 2D, 7034, Trondheim, Norway {\tt\small morten.nissov@ntnu.no}}
}

\begin{document}

\maketitle
\thispagestyle{empty}
\pagestyle{empty}

\begin{abstract}
Enabling autonomous robots to operate robustly in challenging environments is necessary in a future with increased autonomy. For many autonomous systems, estimation and odometry remains a single point of failure, from which it can often be difficult, if not impossible, to recover. As such robust odometry solutions are of key importance. In this work a method for tightly-coupled LiDAR-Radar-Inertial fusion for odometry is proposed, enabling the mitigation of the effects of LiDAR degeneracy by leveraging a complementary perception modality while preserving the accuracy of LiDAR in well-conditioned environments. The proposed approach combines modalities in a factor graph-based windowed smoother with sensor information-specific factor formulations which enable, in the case of degeneracy, partial information to be conveyed to the graph along the non-degenerate axes. The proposed method is evaluated in real-world tests on a flying robot experiencing degraded conditions including geometric self-similarity as well as obscurant occlusion. For the benefit of the community we release the datasets presented: \url{https://github.com/ntnu-arl/lidar_degeneracy_datasets}.
\end{abstract}

\section{Introduction}\label{sec:intro}
Autonomous robots are tasked to navigate in ever more challenging environments with increased requirements for robustness and reliability. To achieve this goal, the hurdles imposed by perceptual degradation must be overcome. Autonomous cars navigating through fog~\cite{bijelic2018benchmark}, aerial robots flying through a self-similar tunnel ~\cite{ebadi2022present,khattak2020complementary} or ground systems mapping a smoke-filled underground mine corridor~\cite{ebadi2022present,CERBERUS_SCIENCE_2022} represent relevant examples. In the quest to enable resilient perception especially for the key task of odometry estimation, identifying appropriate sensing solutions, developing the relevant algorithms to best exploit their data, and combining their strengths through multi-modal fusion are essential steps.

Motivated by the above, this work contributes a novel tightly-coupled, multi-modal odometry framework that exploits the synergy of modern mmWave radars and LiDARs, alongside \acp{imu} to enable resilient estimation in GPS-denied environments that may present geometric self-similarity or dense presence of obscurants such as dust, fog and smoke. With LiDAR methods being particularly sensitive against such conditions~\cite{khattak2020complementary,zhang2015visual,phillips2017dust}, radar fusion is key to overcome degeneracy on the pose estimation problem, while simultaneously supporting accurate velocity estimation. Likewise, limitations of miniaturized radars in the accuracy and density of their spatial point clouds render the sensor applicable for odometry but often with less accuracy compared to LiDAR-based methods~\cite{harlow2023new}. Furthermore, radar data depend on other aspects, such as the material properties of objects as that relates to their \ac{rcs}. Motivated by this we propose a principled strategy to tightly fuse the two modalities considering the complementary nature of their sensing principles. 

\begin{figure}[t!]
    \centering
    \includegraphics[width=\linewidth]{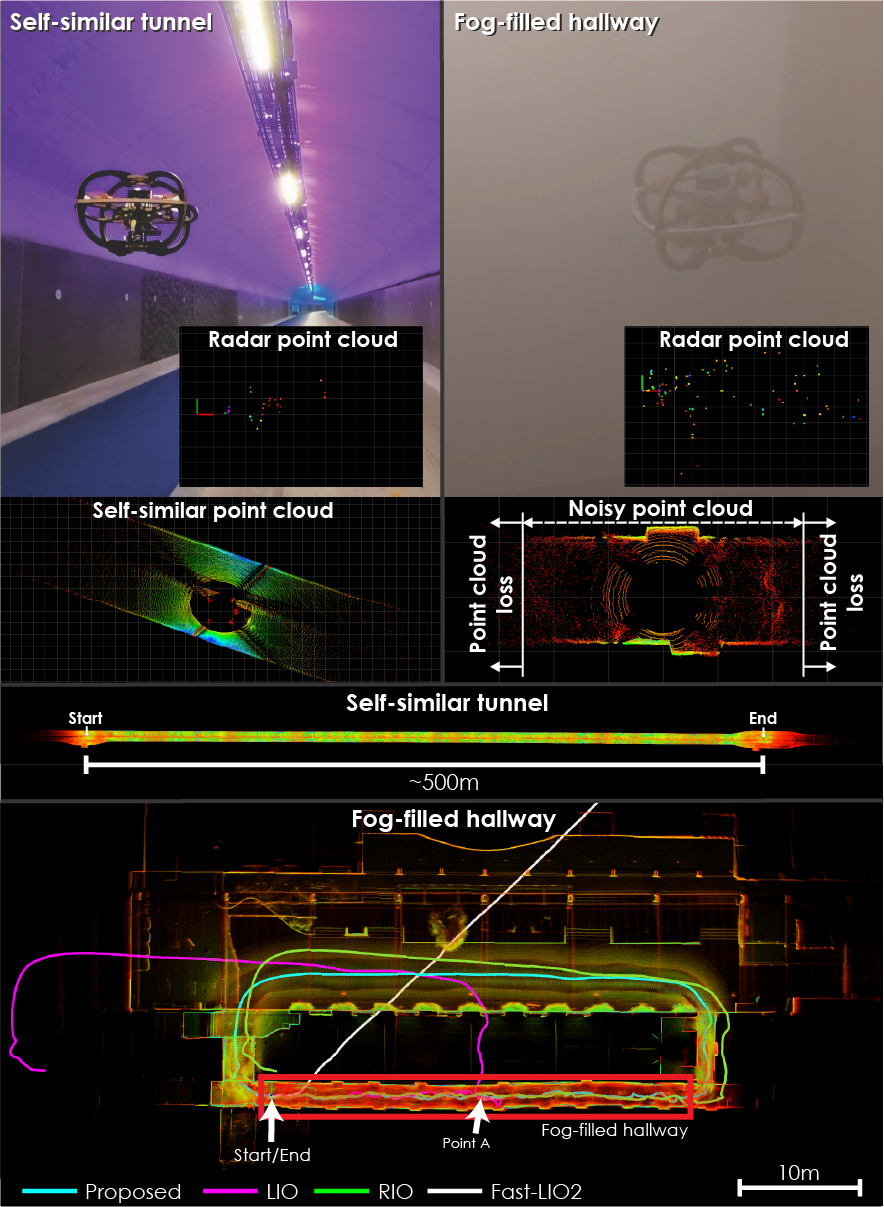}
        \vspace{-5ex}
        \caption{Demonstration of the challenging environments explored in this paper, including a geometrically self-similar tunnel and a building with a fog-filled hallway. The effect of perceptual degradation is shown in the LiDAR point clouds. Radar data is fused to enable resilient odometry.}
    \label{fig:intro}
    \vspace{-5ex}
\end{figure}

The proposed method -- called \ac{method} -- enables robust odometry against perceptually-degraded environments, such as those in \cref{fig:intro}, by tightly-coupled sensor fusion of \ac{imu}, LiDAR, and radar in a sliding window factor graph framework. The fused estimation takes advantage of the strengths of each modality such that 1) divergence is avoided when LiDAR is negatively affected by degeneracy from self-similarity or obscurants since the radar is less affected, and 2) aiding in reducing drift experienced by radar-inertial methods given the higher accuracy of the LiDAR. The integration of the LiDAR through feature factors enables this ability as the graph can receive information regarding the non-degenerate axes even when the LiDAR is experiencing degraded conditions.
The contributions of the proposed method include
\begin{enumerate}[leftmargin=*]
    \item Tightly-coupled LiDAR-radar fusion explored in self-similar and degenerate environments.
    \item Formulation and derivation of factors for direct integration of LiDAR features in the graph with a global map.
    \item Formulation and derivation of a factor for integration of radar-estimated linear velocity in a factor graph for inertial navigation.
\end{enumerate}
To evaluate the performance of the method, a set of representative experimental studies were conducted. Specifically, an aerial robot integrating time-synchronized LiDAR, radar and \ac{imu} sensing was deployed in a) a motion capture arena, b) a long geometrically-self similar bicycle tunnel, and c) a building with a corridor of it being densely filled with fog. 

In the remaining paper, Section~\ref{sec:related} presents related work, followed by the description of the proposed approach in Section~\ref{sec:approach}. Evaluation studies are presented in Section~\ref{sec:evaluation} and conclusions in Section~\ref{sec:conclusions}.


\section{Related Work}\label{sec:related}
The body of work on multi-modal odometry estimation and \ac{slam}, alongside the individual domains of LiDAR-based and radar-based methods relate to this work. Below we overview methods focusing on the fusion of \ac{fmcw} radar with other exteroceptive modalities for odometry and \ac{slam}, while also briefly introducing the diversity of methods on using either only LiDAR or radar as the sole exteroceptive modality. 

\subsection{LiDAR-based SLAM}

LiDAR-based SLAM methods typically solve a registration problem by fitting a candidate point cloud to a reference point cloud. Variations within this domain involve either using the entire point cloud for the registration~\cite{icp,fast-gicp} or selecting descriptive points (features) for computational improvements~\cite{Zhang-RSS-14,legoloam2018shan,compslam,mimosa,zhao2021super,fastlio} while maintaining comparable accuracy. The latter have been widely used in robotics particularly since the platforms that these methods are deployed on may be computationally constrained. A large body of these methods derive from the seminal work of LOAM~\cite{Zhang-RSS-14} which combined high frequency velocity estimation for point cloud deskewing and providing a prior for low frequency high fidelity registration. LOAM derivatives develop on the method in various fashions such as the fusion with other sensors~\cite{liosam2020shan,compslam,mimosa,zhao2021super,fastlio,fastlio2,zhang2015visual} and incorporation of deployment platform information~\cite{legoloam2018shan,pagoloam}. Handling of LiDAR-based measurements in our method is also inspired by LOAM however, is different from it and its derivatives in the formulations of the factors used in the optimization.

\subsection{Radar-based SLAM}

Radar-based odometry and \ac{slam} methods are gaining prominence in robotics, particularly in autonomous driving and small robots, primarily due to weather resilience~\cite{schuster2016landmark,schoen2017real,holder2019real,guan2020through,barnes2020oxford,kramer2022coloradar,lu2020see}. Research includes handling radar distortions~\cite{vivet2013localization}, automotive radar odometry~\cite{aldera2022goes,li2020millimeter,schuster2016landmark,burnett2021radar}, and \ac{slam} solutions~\cite{hong2021radar,hong2020radarslam} including for indoor systems~\cite{marck2013indoor}. Specific developments for small flying robots involve EKF-based radar-inertial odometry methods~\cite{doer2020radar,doer2020ekf,doer2021yaw,doer2021x,michalczyk2022radar}, including tightly-coupled approaches~\cite{michalczyk2022tightly}, and techniques combining radar with \ac{imu} using factor graphs~\cite{kramer2020radar,kramer2021fog}. Recent work demonstrates spinning radar-based localization approximately on par with LiDAR methods~\cite{adolfsson2022lidar}, while the authors in~\cite{burnett2022we} address this question especially having all-weather operations in mind. The survey in~\cite{harlow2023new} overviews the state-of-the-art in mmWave radar applications including for odometry and \ac{slam}. 

\subsection{Radar and LiDAR in Multi-modal SLAM}

A selective niche of works has investigated the potential of fusing LiDAR and radar data for localization and mapping. The work in~\cite{fritsche2018fusing} considered their combination for \ac{slam} involving a mechanical pivoting radar (MPR) either by a) landmarks extracted using LiDAR and MPR, or through b) scan fusion between the two sensors. The authors in~\cite{park2019radar} focused on registering radar measurements on previously built LiDAR maps. The work applied a radar-map refinement step. In a highly relevant manner, the contribution in~\cite{yin2021rall} offers a deep learning-based strategy for radar localization on LiDAR maps. Furthermore, although not itself a multi-modal fusion contribution but focusing on a comparative analysis of LiDAR and radar for \ac{slam}, the work in~\cite{mielle2019comparative} offers insights regarding the expected accuracy and the different challenges faced by the two sensors classes. 



\section{Proposed Approach}\label{sec:approach}
In this section the proposed multi-modal sensor fusion solution capable of resilience against deficiencies in either of the two individual odometry methods by tightly-coupled LiDAR-radar-inertial sensor fusion is proposed. Tightly-coupled in this case refers to the explicit addition of LiDAR features in the graph architecture, reducing the possibility for incorporating errors when making the transition from features derived from degenerate conditions into a single transform estimate, i.e. what can be possible with a more loosely-coupled approach fusing a 6-DoF pose prior~\cite{zhao2021super,nubert2022graph}. This means that the measurement information for each of \ac{imu}, radar, and LiDAR is added to graph through, geometrically- and physically-driven, relationships which are the basis for the factor derivations.

\subsection{Notation}
The notation used in this manuscript is as follows. The variables $x$, $\bm{x}$, $\mathbf{x}$ are scalar, vector, and matrix respectively and the wedge operator $(\cdot)^\wedge$ represents the skew-symmetric matrix in $\mathbb{R}^{3\times 3}$. 
Let the the rotation from $\mathtt{A}$ to $\mathtt{B}$ be ${}_\mathtt{B} \mathbf{R}_\mathtt{A} \in \mathit{SO}(3)$ and the position of frame $\mathtt{A}$ expressed in frame $\mathtt{B}$ be ${}_\mathtt{B} \bm{p}_\mathtt{A} \in\mathbb{R}^{3}$ such that the homogeneous transformation from $\mathtt{A}$ to $\mathtt{B}$ is ${}_\mathtt{B}\mathbf{T}_\mathtt{A} \in \mathit{SE}(3)$.
Furthermore, in this work we use the static world frame ($\mathtt{W}$) and the body-fixed \ac{imu} frame ($\mathtt{I}$), radar frame ($\mathtt{R}$), and LiDAR frame ($\mathtt{L}$). Note, it is assumed that extrinsics between the sensor frames are known a-priori. 

\subsection{State Estimation}
The state space for the proposed method consists of the pose of the \ac{imu} in world frame ${}_\mathtt{W}\mathbf{T}_\mathtt{I} \in \mathit{SE}(3)$, the \ac{imu} linear velocity ${}_\mathtt{W}\bm{v}_\mathtt{I}\in\mathbb{R}^3$, and \ac{imu} biases ${}_\mathtt{I}\bm{b}\in\mathbb{R}^6$ such that the state of a given node in the graph is represented by
\begin{equation}
    \mathbf{x} = \begin{bmatrix}
        {}_\mathtt{W}\mathbf{T}_\mathtt{I} &{}_\mathtt{W}\bm{v}_\mathtt{I} &{}_\mathtt{I}\bm{b}
    \end{bmatrix}
\end{equation}
where the position and orientation of \ac{imu} w.r.t. world (${}_\mathtt{W}\bm{p}_\mathtt{I}$, ${}_\mathtt{W}\mathbf{R}_\mathtt{I}$) make up the pose and ${}_\mathtt{I}\bm{b}$ contains terms for accelerometer $\bm{b}_a$ and gyroscope $\bm{b}_g$.

In the proposed method the estimation problem is solved using a factor graph, performing an incremental optimization over a sliding window to reduce computational cost. 
Having state estimates available at a consistent rate is desirable, and the radar measurement comes with lower latency compared to LiDAR, therefore the radar measurements' timestamps are used to create the graph nodes. Additionally, since the LiDAR will be deskewed, assuming there is at least one radar measurement per LiDAR measurement, the LiDAR can simply be deskewed to the radar measurement timestamp.

\begin{figure}[h]
    \centering
    \includegraphics[width=\linewidth]{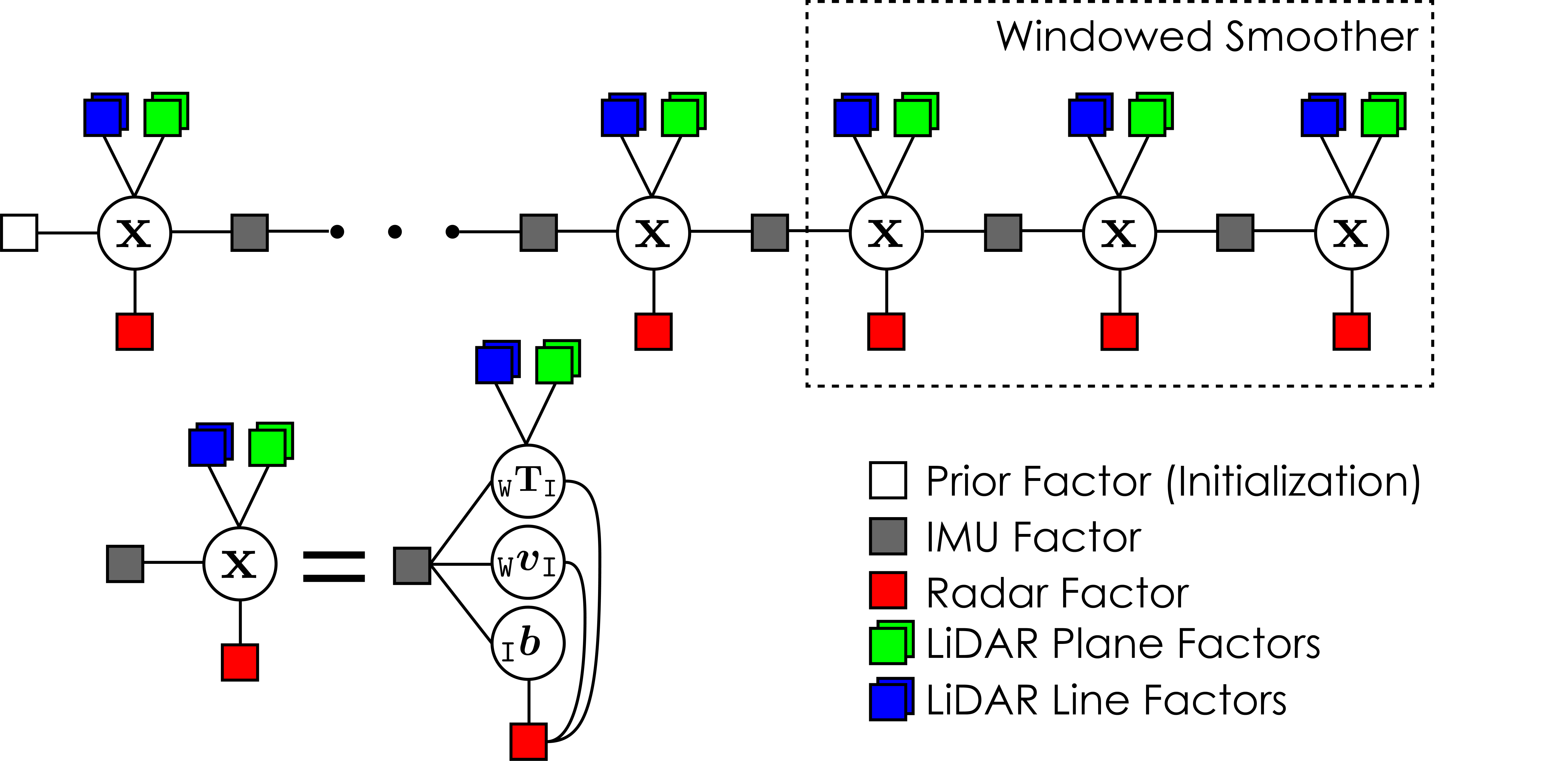}
    \caption{Architecture of the factor graph for the proposed method, including factors created from \ac{imu}, LiDAR, and radar measurements.}
    \label{fig:graph_architecture}
    \vspace{-3ex}
\end{figure}

Let the radar measurement at time $i$ be denoted as $\mathcal{R}_i$. The set of \ac{imu} measurements timestamped between consecutive radar measurements at time $i$ and $j$ is denoted by $\mathcal{I}_{ij}$ and the set of LiDAR feature measurements (consisting of plane and line features described in \cref{sec:approach:lidar}) at time $i$ is $\mathcal{L}_i$.
The set of all radar measurements collected up to time $k$ is $\mathcal{M}_k$ and the set of all measurements collected up to time $k$ is $\mathcal{Z}_k = \{\mathcal{I}_{ij}, \mathcal{L}_i,\mathcal{R}_i\},\ (i,j)\in\mathcal{M}_k$.  Thus, the MAP estimate is
\vspace{-2ex}
\begin{multline}
    \mathcal{X}_{k-l:k}^* = \argmax_{\mathcal{X}_{k-l:k}} p\left( \mathcal{X}_{k-l:k} \vert \mathcal{Z}_{k-l:k} \right) \\\propto p\left( \mathcal{X}_{k-l-1} \right) p\left( \mathcal{Z}_{k-l:k} \vert \mathcal{X}_{k-l:k} \right)
\end{multline}
where $\mathcal{X}_{k-l:k}$ is the windowed set of states from time $k-l$ to $k$.
Assuming Gaussian noise models with zero mean, this can be rewritten into the objective function being optimized by the factor graph such that
\vspace{-2ex}
\begin{multline}
    \mathcal{X}_{k-l:k}^* = \argmin_{\mathcal{X}_{k-l:k}} \Big( \lVert \bm{e}_{k-l-1} \rVert_{\bm\Sigma_0}^2 + \Sigma_{(i,j)\in\mathcal{M}_{k-l:k}} \lVert \bm{e}_{\mathcal{I}_{ij}} \rVert_{\bm\Sigma_{\mathcal{I}}}^2\\ + \Sigma_{i\in\mathcal{M}_{k-l:k}}\lVert \bm{e}_{\mathcal{R}_i} \rVert_{\bm\Sigma_{\mathcal{R}}}^2 + \Sigma_{i\in\mathcal{M}_{k-l:k}}\lVert \bm{e}_{\mathcal{L}_i} \rVert_{\bm\Sigma_{\mathcal{L}}}^2 \Big)
\end{multline}
where $\bm{e}_{k-l-1}$, $\bm{e}_{\mathcal{I}_{ij}}$, $\bm{e}_{\mathcal{R}_i}$, and $\bm{e}_{\mathcal{L}_i}$ are the residuals for the prior and sensor measurement factors and the terms $\bm\Sigma_0$, $\bm\Sigma_{\mathcal{I}}$, $\bm\Sigma_{\mathcal{R}}$, and $\bm\Sigma_{\mathcal{L}}$ represent their covariance matrices, respectively. These residual terms, along with the covariances and Jacobians, define factors which are added to the factor graph such that the final architecture is as shown in \cref{fig:graph_architecture}.
Furthermore, for these factors analytical Jacobians are derived to take advantage of the reduced computational cost compared to online numerical differentiation. The derivations are not provided in this work, see~\cite{solà2021micro} for an introduction to calculus with Lie groups.

\subsubsection{IMU Measurements}
Since \ac{imu} measurements come at a much higher rate than LiDAR or radar, the implementation accumulates \ac{imu} measurements in a buffer, waiting for a radar measurement. Upon receiving a radar measurement a new node is created in the graph and all \ac{imu} measurements from the timestamp of the previous node of the graph up to the the timestamp of the radar measurement are integrated and used to connect the two nodes with a pre-integrated IMU factor~\cite{forster_-manifold_2017} including the following residuals and their covariances
\begin{equation}
    \bm{e}_\mathcal{I} = \begin{bmatrix}
        \bm{e}_{\Delta{}_\mathtt{W}\mathbf{R}_\mathtt{I}}^\top &\bm{e}_{\Delta{}_\mathtt{W}\bm{v}_\mathtt{I}}^\top &\bm{e}_{\Delta{}_\mathtt{W}\bm{p}_\mathtt{I}}^\top
    \end{bmatrix}^\top,\quad \bm\Sigma_\mathcal{I}
\end{equation}
where $\bm{e}_{\Delta{}_\mathtt{W}\mathbf{R}_\mathtt{I}}$, $\bm{e}_{\Delta{}_\mathtt{W}\bm{v}_\mathtt{I}}$, and $\bm{e}_{\Delta{}_\mathtt{W}\bm{p}_\mathtt{I}}$ are the residuals, of the \ac{imu} factor, with respect to the orientation, velocity, and position.

\subsubsection{LiDAR Measurements}\label{sec:approach:lidar}
For each LiDAR point cloud measurement the point cloud is first deskewed. We first propagate the current state to the timestamps of all IMU measurements within the collection period and then interpolate the poses under a constant linear and angular velocity model to obtain the LiDAR pose at the timestamp of each point. Exploiting synchronization between LiDAR and radar and the fact that both run at \SI{10}{\hertz} means the point cloud can be deskewed to the radar timestamp that lies in the collection period of the LiDAR, thus aligning the measurements. The relative pose between the pose at the radar timestamp and the point timestamp is used to deskew the points. After deskewing, features are extracted and used to formulate factors that are then added to the graph. It is important to note here that individual factors based on the features are added instead of a single 6-DoF pose prior (as in \cite{zhao2021super,nubert2022graph}) which would be bound to incorporate incorrect information in degenerate scenarios. This is the case because during LiDAR degeneracy, e.g. in a geometrically self-similar environment, finding correspondences and estimating a single transform is an under-constrained problem and will result in incorrect values along the degenerate axis. By the addition of the individual factors based on features, we ensure that no information is incorporated along the degenerate axis.

Two types of features are considered in the proposed method, namely ``plane'' and ``line'' features. This means that the LiDAR measurements consist of plane feature measurements $\mathcal{L}^p$ and line feature measurements $\mathcal{L}^l$ such that $\mathcal{L} = \mathcal{L}^p\cup\mathcal{L}^l$. These factors are described later in this section.

For each deskewed point cloud, points having high and low curvature are selected as corner and surface features respectively as in~\cite{Zhang-RSS-14}. These two sets are then downsampled and used to find correspondences as 3D lines or planes in their respective local submaps to construct factors to be used in the optimization. The correspondences for the features are found by first finding the $5$ nearest neighbors of the feature point and then using the eigenvalues of the distribution of these neighbors to verify whether they form a 3D line or plane. A k-d tree is built from the submap and used to accelerate this search. The mean of the nearest neighbors is set as a point lying on the corresponding line or plane ${}_\mathtt{W}\bm{j}$ while the eigenvectors of the distribution are used to find the normal to the represented plane ${}_\mathtt{W}\bm{n}$ or the direction vector of the represented line ${}_\mathtt{W}\bm{d}$. 

We now describe the construction of the factors used in the optimization. Note, for both of the LiDAR factors the LiDAR-\ac{imu} extrinsic ${}_\mathtt{I}\mathbf{T}_\mathtt{L}$ is assumed to be known and used to transform point clouds to the \ac{imu}-frame.
\paragraph{LiDAR Point-to-Plane Factor}\label{sec:approach:lidar:point_to_plane}
For a point expressed in the frame of a LiDAR measurement ${}_\mathtt{L} \bm{i}$, the correspondence is parameterized by a point lying on the 3D plane ${}_\mathtt{W}\bm{j}$ and the normal to the plane ${}_\mathtt{W}\bm{n}$. The error is then found as
\begin{equation}
    \bm{e}_{\mathcal{L}^p} = \left(\left({}_\mathtt{W}\bm{i} - {}_\mathtt{W}\bm{j} \right) \cdot {}_\mathtt{W}\bm{n} \right) {}_\mathtt{W}\bm{n}
\end{equation}
with the only non-zero Jacobian taking the form
\begin{equation}
  \frac{\partial \bm{e}_{\mathcal{L}^p}}{\partial {}_\mathtt{W}\mathbf{T}_\mathtt{I}} = {}_\mathtt{W}\bm{n} {}_\mathtt{W}\bm{n}^\top 
  \begin{bmatrix}
      -{}_\mathtt{W}\mathbf{R}_\mathtt{I} {}_\mathtt{I}\bm{i}^\wedge &{}_\mathtt{W}\mathbf{R}_\mathtt{I}
  \end{bmatrix}
\end{equation}

\paragraph{LiDAR Point-to-Line Factor}

For a point ${}_\mathtt{I}\bm{i}$, the correspondence is parameterized by a point ${}_\mathtt{W}\bm{j}$ on the 3D line and its direction vector ${}_\mathtt{W}\bm{d}$. The error takes the form
\begin{equation}
    \bm{e}_{\mathcal{L}^l} = \left( {}_\mathtt{W}\mathbf{T}_\mathtt{I} {}_\mathtt{I}\bm{i} - {}_\mathtt{W}\bm{j} \right) - \left( \left( {}_\mathtt{W}\mathbf{T}_\mathtt{I} {}_\mathtt{I}\bm{i} - {}_\mathtt{W}\bm{j} \right) \cdot {}_\mathtt{W}\bm{d} \right) {}_\mathtt{W}\bm{d} 
\end{equation}

with the only non-zero Jacobians

\begin{equation}
    \frac{\partial \bm{e}_{\mathcal{L}^l}}{\partial {}_\mathtt{W}\mathbf{T}_\mathtt{I}} = \left( \mathbf{I}_3 - {}_\mathtt{W}\bm{d} {}_\mathtt{W}\bm{d}^\top \right)
    \begin{bmatrix}
      -{}_\mathtt{W}\mathbf{R}_\mathtt{I} {}_\mathtt{I}\bm{i}^\wedge &{}_\mathtt{W}\mathbf{R}_\mathtt{I}
    \end{bmatrix}
\end{equation}

After optimization, the resulting state is used to transform the feature clouds into the world frame and update globally maintained maps. For computational efficiency, the maps are maintained at a fixed resolution by downsampling after an update step and local submaps spanning at least the maximum range of the LiDAR are extracted from these maps and used as described above.

\subsubsection{Radar Measurement}
\ac{fmcw} radars create measurements by transmitting high frequency chirps and processing the returns. This process is repeated over an array of antennas to increase the field of view. This results in a point cloud measurement with 3D spatial coordinates, radial speed, and \ac{rcs} per point. The radial speed of a static point is related to the radar linear velocity according to
\begin{equation}\label{eq:approach:radial_speed}
    v_{r} = -{}_{\mathtt{R}}\bar{\bm{r}}^\top {}_{\mathtt{R}}\bm{v}
\end{equation}
where ${}_{\mathtt{R}}\bar{\bm{r}}$ is the point's normalized bearing vector and ${}_{\mathtt{R}}\bm{v}$ is the radar linear velocity. The linear velocity in the $\{\mathtt{R}\}$-frame can be expressed as a function of the state space by
\begin{equation}\label{eq:approach:radar_velocity}
    {}_{\mathtt{R}}\bm{v} = {}_\mathtt{R}\mathbf{R}_\mathtt{I} \left( {}_\mathtt{I}\mathbf{R}_\mathtt{W} {}_\mathtt{W}\bm{v}_\mathtt{I} + \left( {}_\mathtt{I}\bm{\omega} - \bm{b}_{g} \right) \times {}_\mathtt{I}\bm{p}_\mathtt{R} \right)
\end{equation}
where the radar-\ac{imu} extrinsics are given by $\{{}_\mathtt{I}\mathbf{R}_\mathtt{R}, {}_\mathtt{I}\bm{p}_\mathtt{R}\}\in \mathit{SE}(3)$ and ${}_\mathtt{I} \bm\omega$ is the \ac{imu}-frame angular velocity.

Given that a single measurement contains many targets (each composed of a position and doppler measurement) the $\mathbb{R}^3$ velocity can be estimated and made robust against dynamic objects by applying \ac{ransac} on the least squares formulation~\cite{kellner_instantaneous_2013}. The problem can be formulated in a least-squares sense by manipulating and stacking \cref{eq:approach:radial_speed} for each doppler and bearing measurement $v_r^n,\ {}_\mathtt{R}\bar{\bm{r}}^n \ \forall n\in\{1, 2,\ldots, N\}$ for $N$ measurements in a given point cloud
\begin{equation}
\begin{aligned}
    \underbrace{
    \begin{bmatrix}
        -v_{r}^{1}\\
        -v_{r}^{2}\\
        \vdots\\
        -v_{r}^{N}
    \end{bmatrix}
    }_{\bm{v}_r} &= 
    \underbrace{
    \begin{bmatrix}
        \left( {}_\mathtt{R} \bar{\bm{r}}^1 \right)^\top\\
        \left( {}_\mathtt{R} \bar{\bm{r}}^2 \right)^\top\\
        \vdots\\
        \left( {}_\mathtt{R} \bar{\bm{r}}^N \right)^\top
    \end{bmatrix}
    }_{\mathbf{X}}
    \underbrace{
    \begin{bmatrix}
        {}_\mathtt{R} v_x\\
        {}_\mathtt{R} v_y\\
        {}_\mathtt{R} v_z
    \end{bmatrix}
    }_{{}_\mathtt{R} \bm{v}}
\end{aligned}
\vspace{-1ex}
\end{equation}
where $\bm{v}_r$ and $\mathbf{X}$ contain the stacked point measurements from a given point cloud.

\paragraph{Radar Velocity Factor}
Given the $\mathtt{R}$-frame linear velocity estimate ${}_{\mathtt{R}}\tilde{\bm{v}}$, resulting from solving the \ac{ransac} least squares optimization, the radar velocity factor can be added to the graph with an error function derived by combining \cref{eq:approach:radial_speed,eq:approach:radar_velocity}, using estimates from the graph and \ac{imu} measurements of angular velocity
\begin{equation}
    \bm{e}_\mathcal{R} = {}_\mathtt{R}\mathbf{R}_\mathtt{I} \left( {}_\mathtt{I}{\mathbf{R}}_\mathtt{W} {}_\mathtt{W}{\bm{v}}_\mathtt{I} + \left( {}_\mathtt{I}{\bm{\omega}} - {\bm{b}}_{g} \right) \times {}_\mathtt{I}\bm{p}_\mathtt{R} \right) - {}_{\mathtt{R}}\tilde{\bm{v}}
\end{equation}
where the non-zero Jacobians are
\begin{equation}
    \begin{aligned}
        \frac{\partial \bm{e}_\mathcal{R}}{\partial {}_\mathtt{W} \mathbf{R}_\mathtt{I}} &= {}_\mathtt{R}\mathbf{R}_\mathtt{I} \left( {}_\mathtt{I}\mathbf{R}_\mathtt{W} {}_\mathtt{W}\bm{v}_\mathtt{I} \right)^\wedge\\
        \frac{\partial \bm{e}_\mathcal{R}}{\partial {}_\mathtt{W} \bm{v}_\mathtt{I}} &= {}_\mathtt{R}\mathbf{R}_\mathtt{I} {}_\mathtt{I}\mathbf{R}_\mathtt{W}\\
        \frac{\partial \bm{e}_\mathcal{R}}{\partial \bm{b}_g} &= {}_\mathtt{R}\mathbf{R}_\mathtt{I} \left( {}_\mathtt{I}\bm{p}_\mathtt{R} \right)^\wedge
    \end{aligned}
\end{equation}

\paragraph{Uncertainty}
The uncertainty for this factor formulation is calculated according to the equation for estimation the covariance matrix of a least squares solution
\begin{equation}
    \text{cov}({}_{\mathtt{R}}\tilde{\bm{v}}) = \left( \mathbf{X}^\top \mathbf{X} \right)^{-1} \frac{\lVert \bm{v}_r - \mathbf{X}{}_\mathtt{R}\bm{v} \rVert^2}{N-N_{dof}}
\end{equation}
where there are 3 degrees of freedom $N_{dof}$.
Note, this assumes that the uncertainty associated with $\bm{e}_\mathcal{R}$ (i.e. $\mathbf{\Sigma}_\mathcal{R}$) is dominated by ${}_\mathtt{R} \bm{v}$.

The least-squares formulation assumes that uncertainty in the least squares formulation is constrained to be in the dependent variable, where here both dependent and independent contain uncertainty. Specifically, the dependent variable uncertainty arises from uncertainty in the doppler velocity measurement and independent variable uncertainty arises in the azimuth/elevation measurement noise propagating through the bearing vector calculation. Other authors have addressed this~\cite{kellner2013lateral,kellner2014ego}, but the improvement seems to be marginal while it gives rise to high computational costs. Thus, this different approach is not used here.

\subsubsection{Initialization}
An initialization routine, assuming static robot body at start, is added as this increases convergence rate and accuracy during the initial parts of the trajectory. During this time, a few seconds of \ac{imu} data are accumulated for estimating the gravity magnitude (assuming small accelerometer bias), initial roll and pitch, and the gyroscope biases.

\subsection{Implementation Details}
The proposed method is implemented in C++ using the open-source library GTSAM~\cite{gtsam} for factor graph optimization. We utilize the incremental fixed lag smoother provided in the library with a window size of \SI{0.75}{\second}. The feature sets (and corresponding maps) are maintained at a resolution of \SI{0.2}{\meter} resulting in $1000-2000$ factors being added in the larger environments. Additionally, a Huber loss function is used with all LiDAR- and radar-derived factors for robustness to outliers. The proposed method is tuned to run with a consistent output rate of \SI{10}{\hertz} and by leveraging multi-threading it is able to run real-time on an Intel i7-11800H CPU.

Furthermore, in the fog environment, in order to boost the performance a feature heuristic was included to prefer plane features found on the ground. The logic here is that flying low in man-made structures it is more likely that good features can be found on the floor. This is particularly helpful given that the forward-facing radar has the most difficulty with vertical drift. The features on the ground are added following the point-to-plane factor in \cref{sec:approach:lidar}.


\section{Evaluation Studies}\label{sec:evaluation}

The presented multi-modal fusion solution was evaluated on a series of experiments involving an aerial robot flown in a motion capture arena and perceptually degraded environments with geometric self-similarity and dense fog.

\subsection{System Overview}

The proposed method was tested on a variation of the RMF-Owl platform~\cite{rmfOwl} using a VectorNav VN100 \ac{imu} (\SI{200}{\hertz}), Ouster OS0-128 LiDAR (\SI{10}{\hertz}), and a forward-facing Texas Instruments IWR6843AOP-EVM radar (\SI{10}{\hertz}) with chirp configured as seen in \cref{tab:radar_chirp}. The sensors here are synchronized using a custom-designed, microcontroller-based sensor synchronization tool which triggers the \ac{imu} and radar while providing a \SI{10}{\hertz} synchronization signal, originating from a \SI{1}{ppm} real-time clock, to the LiDAR.

\begin{table}[h]
    \centering
    \caption{Configuration parameters for IWR6843AOP-EVM chirp}
    \label{tab:radar_chirp}
    \vspace{-2ex}
    \begin{tabular}{ll}
        \toprule
        Parameter       &Value\\
        \midrule
        Start frequency &\SI{60}{\giga\hertz}\\
        Bandwidth       &\SI{1911.273}{\mega\hertz}\\
        Maximum range   &\SI{15.999}{\meter}\\
        Maximum doppler   &\SI{3.995}{\meter\per\second}\\
        Range resolution    &\SI{0.0785}{\meter}\\
        Doppler velocity resolution &\SI{0.133}{\meter\per\second}\\
        Azimuth/Elevation resolution  &\SI{29}{\degree}\\
        \bottomrule
    \end{tabular}
    \vspace{-4ex}
\end{table}

\subsection{Experiment 1: Motion Capture Arena}
\begin{figure}[h]
    \centering
    \includegraphics[width=\linewidth]{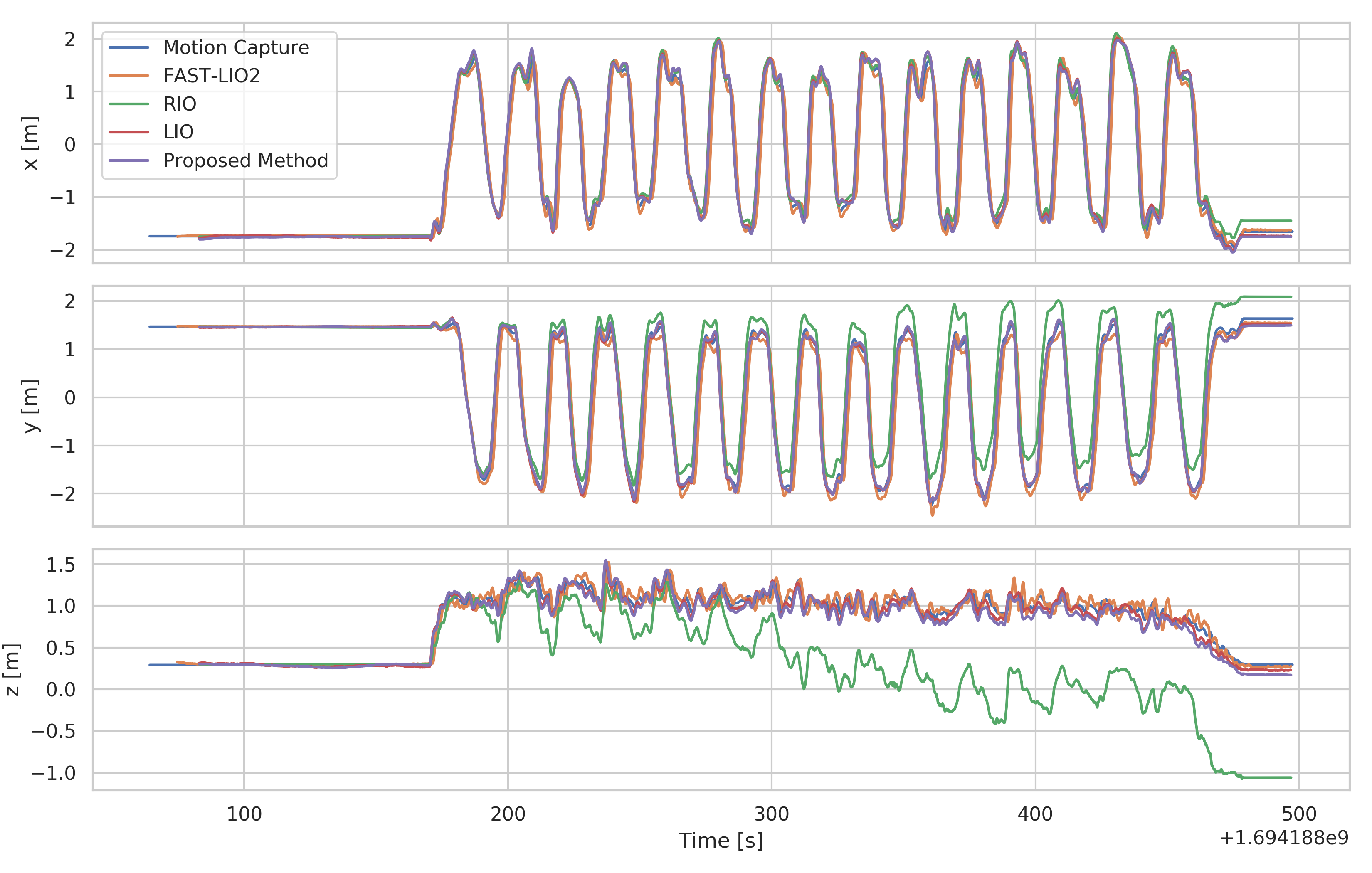}
    \vspace{-4ex}
    \caption{Motion capture results comparing FAST-LIO2, the proposed method, and its versions without LiDAR (RIO) or radar (LIO).}
    \label{fig:mocap_xyz}
    \vspace{-2ex}
\end{figure}

\begin{figure*}[h]
    \centering
    \includegraphics[width=\linewidth]{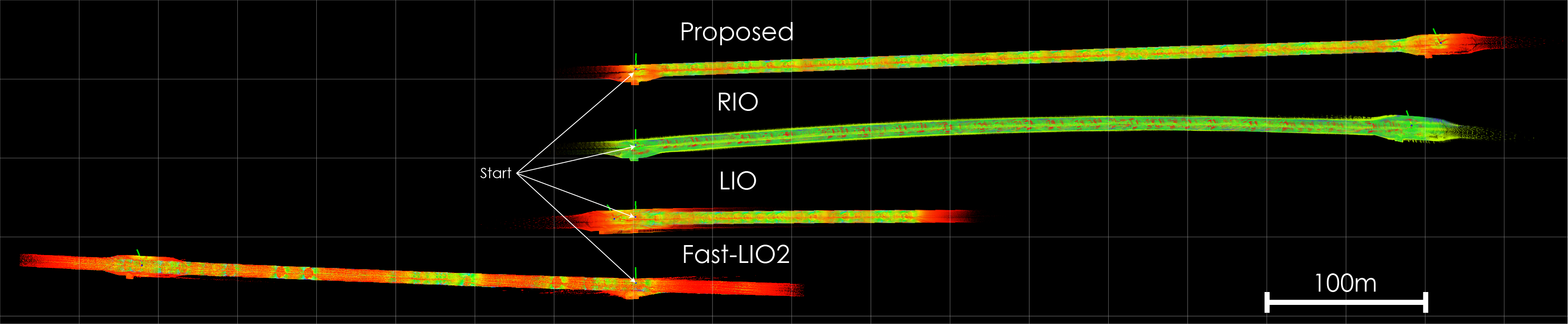}
    \vspace{-4ex}
    \caption{Trajectories of the proposed method, the proposed method without LiDAR (RIO), the proposed method without radar (LIO), and FAST-LIO2 in the geometrically self-similar section of the Fyllingsdalen bicycle tunnel. Note FAST-LIO2 diverges backward after entering the self-similar region.}
    \label{fig:bergen_tunnel}
    \vspace{-2ex}
\end{figure*}

\begin{figure*}[h]
    \centering
    \includegraphics[width=\linewidth]{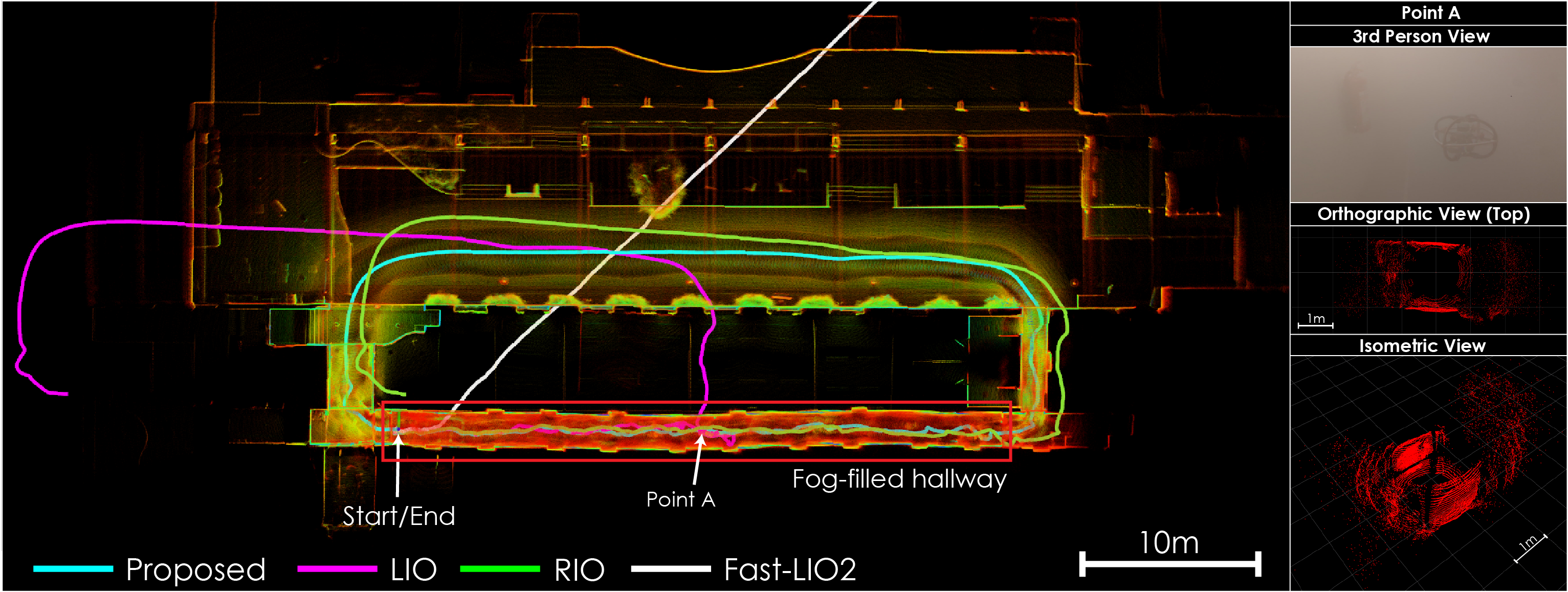}
    \vspace{-4ex}
    \caption{Left: Trajectories of the proposed method, the proposed method without radar (LIO), the proposed method without LiDAR (RIO), and FAST-LIO2 in the NTNU fog environment. Note FAST-LIO2 almost immediately diverges due to fog. Right: Third person camera view and views of the raw pointcloud of the LiDAR at Point A in the fog-filled hallway}
    \label{fig:ntnu_fog}
    \vspace{-3ex}
\end{figure*}

Flight datasets were captured in the motion capture arena at NTNU. This is used here as evidence for claims regarding the validity of the least-squares estimation of 3-vector velocity from a radar point cloud measurement. 
The motion capture velocity estimate was used to calculate the radar velocity measurement error. Based off of the error statistics in
\cref{tab:experiments:mocap:velocity_error}, we conclude that the measurement quality is sufficient given the low error and standard deviation.

Furthermore, position estimates of the proposed method are compared against FAST-LIO2~\cite{fastlio2}, a state-of-the-art open-source LiDAR-inertial solution, as well as versions of our method with the LiDAR or radar disabled denoted as RIO and LIO respectively. The resulting trajectories are shown in \cref{fig:mocap_xyz} and the \ac{ape} in \cref{tab:mocap:stats}.
From this one can see the inevitable drift and reduced accuracy of the radar compared with the LiDAR-based methods, as well as the ability of the proposed method to incorporate information from the LiDAR to reduce error and produce results comparable to FAST-LIO2. Note, although LIO has a slightly lower \ac{ape} than \ac{method}, the difference is small enough that it is considered to be negligible. Note also, the RIO performance in motion capture is indicative of the expected performance: drift in yaw (resulting here in $y$-drift) and drift in $z$, where the latter tends to be greater.

\begin{table}[h]
    \centering
    \caption{Statistics of radar velocity estimate error}    \label{tab:experiments:mocap:velocity_error}
    \vspace{-2ex}
    \begin{tabular}{ccc}
        \toprule
        \multicolumn{3}{c}{Error: mean (standard deviation) [\si{\meter\per\second}]}\\
        ${}_\mathtt{R}\bm{v}_x$   &${}_\mathtt{R}\bm{v}_y$  &${}_\mathtt{R}\bm{v}_z$\\
        \midrule
        0.005	(0.048)	&0.002	(0.039)	&0.010	(0.059)	\\
        \bottomrule
    \end{tabular}
\end{table}
\vspace{-4ex}

\begin{table}[h]
    \centering
    \caption{\ac{ape} for different methods on  motion capture dataset}
    \label{tab:mocap:stats}
    \vspace{-2ex}
    \begin{tabular}{ccccc}
        \toprule
         &FAST-LIO2  &RIO  &LIO  &\ac{method}\\
        \midrule
         \ac{ape}~[\si{\meter}]  &0.268 &0.865 &0.239 &0.242\\
        \bottomrule
    \end{tabular}
    \vspace{-3ex}
\end{table}

\subsection{Experiment 2: Geometrically Self-similar Tunnel}\label{sec:experiment2}
As a second experiment we collect a dataset of the robot being manually piloted in an \SI{8}{\meter}-wide, \SI{500}{\meter}-long straight section of the Fyllingsdalen bicycle tunnel in Bergen, Norway. The robot takes off near a rest area in the tunnel which is non self-similar, flies through a self-similar section and lands in the next non-self-similar rest area. The reconstructed point cloud by the proposed method and the estimated trajectories are shown in \cref{fig:bergen_tunnel}.

Both FAST-LIO2 and LIO can be seen to diverge almost immediately after entering the self-similar section due to the degeneracy in the point cloud data. 
Both RIO and \ac{method} successfully complete the tunnel, however, due to the factors incorporated from the LiDAR that constrain the yaw of the estimation, the result of \ac{method} is seen to have less drift in its yaw.
Note that, although minimal and not easily distinguishable in the maps, there remains some z-drift present in both the RIO and \ac{method} solutions due to the fact that there were not many valid LiDAR features on the ceiling or floor in this environment, resulting in little information in this direction.

\subsection{Experiment 3: Fog-filled hallway}
Finally, we collect a dataset of the robot flying in a loop (approx \SI{100}{\meter}) in a building at NTNU in Trondheim, Norway. The first portion of this trajectory involves a narrow hallway before moving out to a more open area. The narrow hallway is intentionally filled with dense theatrical fog to provide a degenerate environment for the LiDAR. Afterwards the robot exits the narrow hallway, proceeds though the open area where LiDAR has no difficulties, and returns to the starting position. The point cloud, reconstructed by the proposed method, as well as the trajectory can be seen in \cref{fig:ntnu_fog}.
Particularly, the fog severely reduces the effective range of the sensor so that there are only a few valid points on the walls and the floor within \SI{2}{\meter}. FAST-LIO2 can be seen to break immediately after entering the fog due to the large amount of noise in and extremely short range of the point cloud. LIO fares a little better, but underestimates the length of the hallway significantly as the features it detects do not constrain it along the direction of the hallway. RIO is unaffected by the fog, but drifts in its altitude and yaw. Due to the fusion of information only along informative directions, \ac{method} progresses through the hallway with minimal yaw or altitude drift while also providing a better estimate of the length of the hallway as evaluated by the return of the estimate to within \SI{1}{\meter} of the starting position.


\section{Conclusions}\label{sec:conclusions}
This paper presented \ac{method}, a tightly-integrated LiDAR-Radar-\ac{imu} fusion odometry method. The proposed method combines the advantages of LiDAR and radar odometry methods in a tightly-coupled fashion using windowed optimization in a factor graph architecture with novel factor formulations tailored to the sensing modalities. Doing so enables odometry to be robust to typical LiDAR degenerate conditions as well as maintaining the expected performance of LiDAR-inertial methods when in decent conditions. Real world tests conducted onboard flying robots demonstrate the ability to operate in both geometrically self-similar and visual-obscurant filled environments.


\section*{ACKNOWLEDGMENT}
We would like to thank the Vestland Fylkeskommune for providing access to the Fyllingsdalen sykkeltunnel.

\bibliographystyle{IEEEtran}
\bibliography{BIB/Radar-Literature,BIB/LiDAR-Literature,BIB/multi-modal,BIB/General}

\end{document}